# Forecasting blood sugar levels in Diabetes with univariate algorithms

Ignacio Rodríguez[1]

Abstract. AI procedures joined with wearable gadgets can convey exact transient blood glucose level forecast models. Also, such models can learn customized glucose-insulin elements dependent on the sensor information gathered by observing a few parts of the physiological condition and every day movement of a person. Up to this point, the predominant methodology for creating information driven forecast models was to gather "however much information as could be expected" to help doctors and patients ideally change treatment. The goal of this work was to examine the base information assortment, volume, and speed needed to accomplish exact individual driven diminutive term expectation models. We built up a progression of these models utilizing distinctive AI time arrangement guaging strategies that are appropriate for execution inside a wearable processor. We completed a broad aloof patient checking concentrate in genuine conditions to fabricate a strong informational collection. The examination included a subset of type-1 diabetic subjects wearing a glimmer glucose checking framework. We directed a relative quantitative assessment of the presentation of the created information driven expectation models and comparing AI methods. Our outcomes show that precise momentary forecast can be accomplished by just checking interstitial glucose information over a brief timeframe and utilizing a low examining recurrence. The models created can anticipate glucose levels inside a 15-minute skyline with a normal mistake as low as 15.43 mg/dL utilizing just 24 memorable qualities gathered inside a time of 6 hours, and by expanding the inspecting recurrence to incorporate 72 qualities, the normal blunder is limited to 10.15 mg/dL. Our forecast models are reasonable for execution inside a wearable gadget, requiring the base equipment necessities while simultaneously accomplishing high expectation precision.

## 1. Introduction

Having drafted a diagram about equipment and correspondences added to factors and their significance, quite possibly the main assignments to oversee DM1 is to envision to circumstances like hypo or hyperglycaemia. Various endeavors have been made to create models for anticipating blood glucose levels in individuals with diabetes. Today, the most all around read approach for blood glucose forecast depends on point by point physiological models that attempt to catch the elements of glucose-pertinent factors inside various frameworks in the body. All the more as of late, another methodology has been proposed to address this issue by applying AI procedures [1]. The vital idea driving this methodology is that there exist monotonous cycles in glucose-insulin elements, e.g., previously/after suppers, previously/after sleep time, and unsurprising insulin affectability changes for the duration of the day in view of circadian varieties in chemical levels. Striking investigations have assessed designs by season of day, in any event, indicating the presence of changeability as per the day of the week [2]. A few information driven models have been created to investigate the redundant idea of glucose-insulin elements, utilizing assorted procedures for time arrangement estimating, for example, autoregressive (AR), motivation reaction (IR), autoregressive exogenous info (ARX), autoregressive moving normal exogenous information (ARMAX), and autoregressive incorporated moving normal (ARIMA). For instance, Estrada et al. built up an ARX model that depends on blood glucose levels and insulin measurements to estimate future blood glucose levels inside a 45-minute forecast skyline [3]. Different methodologies, as performed by Nuryani et al., investigated the utilization of help vector machines (SVM) to anticipate hypoglycemia utilizing electrocardiograms (ECG) notwithstanding blood glucose levels and insulin infusions [4]. Marling et al. utilized the SVM models, joining information accumulated from wearable movement trackers, checking pulse, galvanic skin reaction, and skin/air temperatures [5].

---
[1] I.R: is with University of Málaga (Spain)

Another class of experimental models utilizes counterfeit neural organizations (ANNs) [6] to get familiar with the connection among past and future blood glucose levels, additionally contemplating other information sources. For instance, Pappada et al. built up a methodology that consolidated the fingerstick strategy to gauge blood glucose levels with insulin doses, dinners, and here and there way of life conditions and feelings [7]. Zecchin et al. built up an indicator dependent on a neural organization (NN) model and a first-request polynomial extrapolation calculation that joins past CGM sensor readings with information on sugar consumption [8]. Tragically, NN-based models require an enormous volume of information to appropriately ascertain the inner boundaries of the organizations and stay away from overfitting. Also, most of these examinations surveyed the exhibition of the models either utilizing reenacted datasets or certifiable information assortment missions of exceptionally restricted span. It is hence essential to ensure that information utilized for the preparation of a particularly experimental model is by configuration long haul and includes genuine people.

Along these lines, we can discover numerous instances of glucose determining including past glucose esteems, yet additionally utilizing different highlights, for example, insulin routine, suppers, work out, and so on The chance of determining utilizing just past glucose esteems has been considered, and this could have a few advantages: depending just on one gadget (CGM), evading human blunder because of subjectivity, and streamlining the figuring cycle of the calculation. With this, we search for the chance of doing this errand locally, for example in a cell phone, which could be helpful under certain conditions (shaky web association, far off areas).Although demonstrating and determining glycaemia utilizing distributed computing present numerous favorable circumstances, we will assess on the off chance that it very well may be conceivable to make an interpretation of this twofold errand to an obliged gadget (rearranging the work by utilizing univariate approaches) to utilize the restricted computational force existing in little gadgets under inaccessibility of the cloud assets.

The idea of creating information driven expectation models depending exclusively on CGM advancements has been concentrated in the past [9]. Distinctive AI methods have been utilized to create information driven expectation models that can be utilized for early hypoglycemic/hyperglycemic alerts and for shutting the glucose guideline circle with an insulin siphon. For instance, Sparacino et al. assessed an AR-based model inside a medical clinic climate including 28 DM1 patients for 48 hours [10]. Eren-Oruklu et al. built up a model utilizing ARIMA dependent on information gathered from observing 22 DM1 patients for 48 hours. Hamdi et al. checked 12 patients to build up a model joining support vector relapse and a differential advancement calculation [11].

Regardless, to the best of the creators' information, an appropriate and complete correlation between glucose forecast calculations has not been done already. Best case scenario, in certain works a couple of techniques were looked at, utilizing glycemia yet in addition (contingent upon the paper) a couple of different variables, accepting a few boundaries, and it is preposterous to expect to remove in general ends. Likewise, the root of the information was genuinely not quite the same as one examination to another, so building up a correlation among the strategies concentrated in various works turns into an undertaking just feasible in an emotional manner.

## 2. The glycaemia forecasting task

As we referenced, one of the principle attributes of a DM1 the executives structure is a guaging blood glucose model. To start with an improved issue, our objective was to create dependable forecast models that can gauge the future glucose level with high precision dependent on current and past information gathered from the FGM sensor. We created persistent driven models by utilizing diverse AI techniques to catch the properties of the blood glucose time arrangement of every individual patient [12]. Subsequently, for every patient, we produced a different forecast model.

The information gave by the FGM sensor [13] were inspected with an example recurrence (SF) of 1 estimation like clockwork, 10 minutes or 15 minutes. In this way, the SF controlled the speed of the information that we are considering. The qualities inspected were utilized to make a past sliding window (PSW) including noteworthy qualities changing from 3 to 36 hours. The PSW controlled the volume of the information the model utilized for the expectation. Given the sliding window information, the model ceaselessly anticipated the glucose levels at preset forecast skylines (PH) at 15, 30, 45, and an hour ahead from right now [14]. Figure 1 gives a graphical portrayal of the fracture of the information gathered into windows that are utilized as contribution for the patient-driven expectation model.

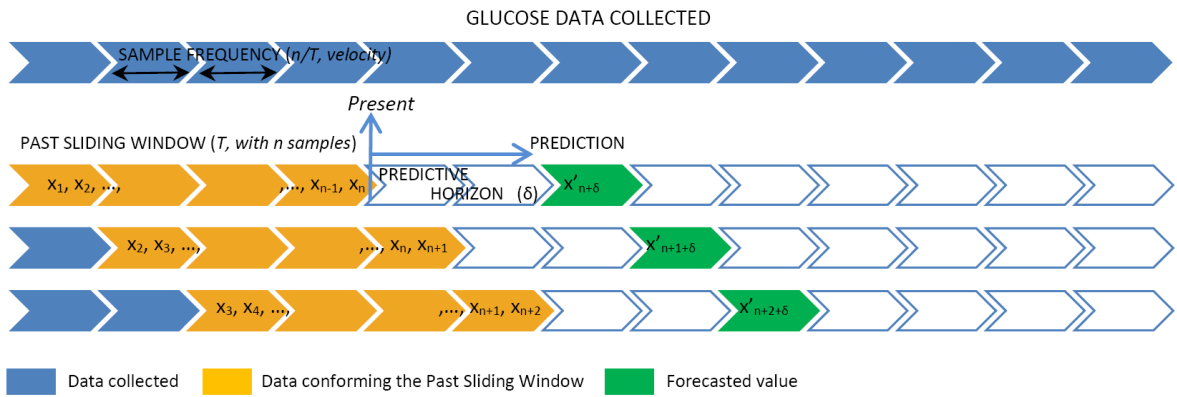

Figure 1. Time series analysis and cross-validation with slide-window.

In more detail, at a given time t the information gathered from the FGM sensor over the past period T were utilized to create the preparation set of n information focuses $\{x_i\}_{i=1}^n$, where $x_i \in R$ is every individual information point got from the FGM sensor. We state that n portrays the volume of the informational index and n/T the speed of the information. Given this preparation set, the objective of the forecast model is to surmised the genuine basic planning precisely enough with the end goal that it can anticipate the following δ esteem, that is, the yield esteem y(t + δ) ∈ R for that time arrangement. We state that δ is the forecast skyline.

The sliding window functions as follows. When another FGM esteem is gotten, the preparation set is revamped by eliminating the most seasoned perception (x1), moving all the qualities 1 situation up (i.e., xi becomes xi−1) lastly adding the new worth got as the freshest one (xn). Along these lines, the dataset size and requesting of the perceptions is constantly safeguarded.

The strategies considered in this first methodology for glucose level forecast are the accompanying:

- Autoregressive integrated moving average (ARIMA).
- Random forest (RF).
- Support vector machines (SVM)[15].

We start by investigating the accomplished presentation in foreseeing future blood glucose levels when we utilize just recorded information on the blood glucose levels. Table 3 shows the outcomes got regarding root mean square blunder (RMSE) between the anticipated qualities and the noticed qualities. We initially see that for all the patient models created, the mistake of the expectation is more prominent as the forecast skyline (PH) builds, which appears to be sensible since the information gathered is continuously further from the forecast. Regardless, as a rule, for the three unique strategies used to build up the patient models, adequate forecasts are accomplished for PHs of 15 minutes and 30 minutes, with normal RMSEs essentially more modest than 20 mg/dL for ARIMA and RF, and more modest than 28 mg/dL for SVM. Similar approaches have been studied before successfully [16][17].

We proceed at Table 1 by investigating the significance of recorded information volume on the expectation precision of future glycemic levels. The size of the window of past information permits the model to catch the transient structure in the time arrangement. One expects that expanding the volume of recorded information will improve the expectation exactness. Consequently, we planned the principal tests via preparing the patient models utilizing distinctive past sliding window (PSW) sizes that contained the perceptions of the previous 3, 6, 12, 24, and 36 hours. In this investigation, the information speed is fixed to 1 example like clockwork and an expectation skyline of 15, 30, 45, and an hour.

Investigating the volume of verifiable information expected to accomplish a precise forecast, as communicated by the distinctive PSW sizes utilized, the outcomes accomplished demonstrate that all the patient models created have a diminished RMSE with a window size of 6 hours.

Exploratory proof shows that paying little mind to the strategy used to build up the models, and for all the forecast skylines, there is a breaking point moving in reverse (6 hours) with regards to considering past information to improve precision of the expectation. Past works have presented this significant degree, connecting the possibility of circadian cycles [18] and the spaces of morning/evening/night.

Table 1. Root mean square error (RMSE in mg/dL) for different past sliding window sizes (PSW) and for different predictive horizon (PH) using a fixed sampling frequency (SF = 5 minutes)

| $SF = 5$ min | | | | | |
|---|---|---|---|---|---|
| RMSE (mg/dl) | | $PH$=15 | 30 | 45 | 60 min |
| ARIMA | 3 | 11.64 | 17.62 | 23.21 | 28.17 |
| | 6 | 11.53 | 17.31 | 22.75 | 27.60 |
| $PSW$ (hours)= | 12 | 13.00 | 18.64 | 23.92 | 28.64 |
| | 24 | 13.93 | 19.45 | 24.55 | 29.05 |
| | 36 | 14.78 | 20.21 | 25.24 | 29.68 |
| RF | 3 | 10.59 | 15.07 | 19.04 | 22.56 |
| | 6 | 10.15 | 14.63 | 18.60 | 22.12 |
| $PSW$ (hours)= | 12 | 11.18 | 15.66 | 19.64 | 23.17 |
| | 24 | 11.65 | 16.11 | 20.06 | 23.56 |
| | 36 | 11.98 | 16.43 | 20.38 | 23.89 |
| SVM | 3 | 18.14 | 22.55 | 26.46 | 29.74 |
| | 6 | 17.65 | 20.82 | 23.74 | 26.36 |
| $PSW$ (hours)= | 12 | 19.73 | 22.57 | 25.17 | 27.48 |
| | 24 | 21.45 | 26.73 | 30.71 | 22.68 |
| | 36 | 23.08 | 27.90 | 31.30 | 33.90 |

Table 2. Root mean square error (RMSE in mg/dL) for different sample frequencies (SF) and for different predictive horizon (PH) using a fixed past sliding window size (PSW = 6 hours).

| $PSW = 6$ hour | | | | | |
|---|---|---|---|---|---|
| RMSE (mg/dl) | | $PH$=15 | 30 | 45 | 60 min |
| ARIMA | 5 | 11.53 | 17.31 | 22.75 | 27.60 |
| $SF$ (min)= | 10 | 13.14 | 20.13 | 23.64 | 29.81 |
| | 15 | 15.08 | 21.10 | 26.32 | 30.82 |
| RF | 5 | 10.15 | 14.63 | 18.60 | 22.12 |
| $SF$ (min)= | 10 | 11.65 | 17.37 | 19.84 | 24.26 |
| | 15 | 15.43 | 19.41 | 22.87 | 25.92 |
| SVM | 5 | 17.65 | 20.82 | 23.74 | 26.36 |
| $SF$ (min)= | 10 | 19.90 | 23.97 | 25.73 | 28.90 |
| | 15 | 23.26 | 26.03 | 28.44 | 30.57 |

Investigating the presentation accomplished by the three distinct strategies used to build up the patient models, the outcomes show that RF technique is the most exact one, accomplishing better forecasts with all the PSWs and for all the PHs considered.

We currently continue by investigating the impact of the speed of the information gathered from the GCM sensor on the expectation capacity of the patient models. In the past arrangement of tests, the information speed, (i.e., the SF), was set to 1 example at regular intervals. In this arrangement of tests, we assessed the impact of utilizing distinctive inspecting frequencies somewhere in the range of 5 and 15 minutes on the accomplished RMSE. Table 2 portrays the outcomes got regarding RMSE between the anticipated qualities and the noticed qualities when the past sliding window is fixed to 6 hours, as this was the ideal worth distinguished in the past investigation. It is apparent that the higher the SF, the higher the subsequent RMSE, paying little heed to the PH. Nonetheless, as the PH builds, the impact of the SF is greater: in a PH of 15 minutes, while expanding the SF to 10 minutes, RMSE expanded by half; yet with a similar change in SF, in a PH of an hour, RMSE expanded distinctly around 10% to 12%. We presume that the choice of lessening SF relies upon how far we need to foresee. An intriguing point is that, just by changing SF from 5 minutes to 10 minutes, satisfactory RMSEs are as yet accomplished. We additionally saw that the patient models that depend on the SVM technique arrived at an exceptionally high RMSE when SF is equivalent to 15 minutes. We hence reason that models dependent on the SVM [19] strategy should utilize a SF of 5 minutes [20].

Similarly as with the past trial, we noticed again that the patient models created utilizing the arbitrary timberland strategy accomplish a superior presentation for each extraordinary SF considered [21].

## 3. Conclusions

The proliferation of ICT solutions [22] (IoT among them[23]) represents new opportunities for the development of new intelligent services, contributing to more efficient and sustainable management of many different aspects of our society. Helping people in the field of health is a remarking issue to be taken into account. Being in these days Diabetes Mellitus reaching heights increasingly, it is critical to lead our efforts to apply ICT to help people with diabetes. Waves propagation also introduces novelties in this field that can be taken into account in future works [24][25][26].

Given the constant evolution of technologies and protocols that are emerging for the IoT [27], in the scope of this work, we have proposed the definition of an architectural framework that abstracts from the underlying technologies for managing different requirements of a person with diabetes. The proposed design enables connectivity between patient and caregivers, and allows the patient to take advantage of optimized solutions [28]. In addition, within this framework, we have introduced novel features to be taken into account, and improved motivation of the patient [29].

Another contribution of this thesis is the development of a proper data set, which is a novelty itself, since it engaged a quantity of subjects, for such a period of time, and with a completed monitoring, that makes this novel data set, to the best of the authors' knowledge, as the most complete until now [30]. This data set will open the door to future works.

At the end, the study of forecasting glycaemia in a simple and affordable way was another novelty exposed in this thesis, considering the idea of performing data modeling and prediction in a constrained device. For this purpose, we used simple and well-known univariate algorithms. Experimental evidences show that such forecasting is possible under some circumstances.

All in all, this work describes novel ways to continue developing paths to properly manage diabetes by using technologies [31] and machine learning techniques, making patients' lives easier while at the same time lowering the risks associated to this illness.

For such purposes, in addition to the efforts that have been taken place in many universities and research centers in the world, we have humbly contributed with the results presented in this work.